\documentclass[journal]{IEEEtran}

\usepackage[compress]{cite}
\usepackage{amsthm, amsmath, stackrel}
\usepackage{amssymb}
\usepackage{graphicx}
\usepackage{mathrsfs}
\usepackage{mathtools}
\usepackage{savesym}
\usepackage{cuted}
\usepackage{nccmath}
\usepackage{amsfonts}
\usepackage{epstopdf}
\usepackage{lipsum}
\usepackage{float}
\usepackage{stfloats}
\usepackage{url}
\usepackage{subcaption}
\usepackage{algorithm}
\usepackage{algpseudocode}
\usepackage{algcompatible}
\usepackage[super]{nth}
\usepackage[english]{babel}
\usepackage{multirow}
\usepackage{makecell}
\usepackage{soul}
\usepackage[dvipsnames]{xcolor}

\providecommand{\U}[1]{\protect\rule{.1in}{.1in}}

\makeatother

\begin{document}

\title{An Explainable Regression Framework for Predicting Remaining Useful Life of Machines}

\author{Talhat Khan\IEEEauthorrefmark{1},
Kashif Ahmad\IEEEauthorrefmark{2},
Jebran Khan\IEEEauthorrefmark{3}, Imran Khan\IEEEauthorrefmark{1},  Nasir Ahmad \IEEEauthorrefmark{1} \\
\IEEEauthorblockA{\IEEEauthorrefmark{1} Department of Computer Systems Engineering, University of Engineering and Technology, Peshawar, Pakistan. \\
\{n.ahmad@uetpeshawar.edu.pk\}} \\
\IEEEauthorblockA{\IEEEauthorrefmark{2} Department of Computer Science, Munster Technological University, Cork, Ireland, kashif.ahmad@mtu.ie} \\
\IEEEauthorblockA{\IEEEauthorrefmark{3} Department of Artificial Intelligence, AJOU University, Suwon, South Korea, jebran@ajou.ac.kr} \\
}

\maketitle

\begin{abstract}
Prediction of a machine's Remaining Useful Life (RUL) is one of the key tasks in predictive maintenance. The task is treated as a regression problem where Machine Learning (ML) algorithms are used to predict the RUL of machine components. These ML algorithms are generally used as a black box with a total focus on the performance without identifying the potential causes behind the algorithms' decisions and their working mechanism. We believe, the performance (in terms of Mean Squared Error (MSE), etc.,) alone is not enough to build the stakeholders’ trust in ML prediction rather more insights on the causes behind the predictions are needed. To this aim, in this paper, we explore the potential of Explainable AI (XAI) techniques by proposing an explainable regression framework for the prediction of machines' RUL. We also evaluate several ML algorithms including classical and Neural Networks (NNs) based solutions for the task. For the explanations, we rely on two model agnostic XAI methods namely Local Interpretable Model-Agnostic Explanations (LIME) and Shapley Additive Explanations (SHAP). We believe, this work will provide a baseline for future research in the domain.

\end{abstract}

\begin{IEEEkeywords}
Explainability, Interpretability, Predictive Maintenance, Regression, Remaining Useful Life, LIME, SHAP.
\end{IEEEkeywords}
\IEEEpeerreviewmaketitle

\section{Introduction}
\label{sec:introduction}
In the modern world, the scope of industries has expanded a lot. These days industries are generally equipped with a large number of modern machines resulting in a significant increase in production. However, the performance of these machines may degrade over time if proper care is not taken, thus, they need a continuous monitoring and maintenance process. Maintenance of machines in industries is a tedious and time-consuming process and generally needs to take different factors into account. However, thanks to the recent advancement in technology, industry 4.0 have opened new opportunities for predictive maintenance \cite{jagatheesaperumal2021duo}.  

Predictive maintenance is one of the key aspects of modern industries especially after the fourth revolution of industry. It allows for monitoring, analyzing, and determining the condition of machine components for early detection of potential faults. The process generally involves data acquisition, data processing, and making intelligent decisions on the basis of the collected data to improve and optimize maintenance processes. Predictive maintenance generally involves different tasks. One of these tasks is the prediction of remaining useful life (RUL) of machines installed in a factory.   

Thanks to the recent advancement in Machine Learning (ML) and sensor technology, it is possible to automate the prediction of RUL of machines by training ML algorithms on the data collected through a diversified set of sensors installed in the machines. The literature already reports the effectiveness of a wide range of ML algorithms in predictive maintenance in general and in predicting the RUL of machines in particular. However, these ML algorithms are used as a black box with total focus on their performance (i.e., accuracy and Mean Squared Error, etc.,) without providing any insights on the working mechanism and cause behind the decisions of these algorithms. In such a critical application, accuracy/MSE alone is not enough to build the stakeholders' trust in ML prediction rather more insights on the causes behind the predictions are needed \cite{ahmad2022developing}. 

In this paper, we propose an explainable ML framework incorporating a couple of model agnostic explainable AI techniques for the prediction of the RUL of machines. The proposed framework provides insights into the ML model's predictions allowing the stakeholders to analyze the main causes of machine degradation. These insights not only build stockholders' trust in the framework but also allow them to tune the model for better predictive performance.

The key contributions of the work can be summarized as follows:

\begin{itemize}
    \item We propose an explainable framework incorporating multiple regression and explainability methods for predicting the RUL of machines.
    \item We also evaluate several regression methods including multiple classical and Neural Networks (NNs) based techniques. Moreover, we analyze the results of two model agnostic methods for the explanation of our regression algorithms.
    \item In extensive experimental setup, we evaluate the potential and applicability of explainable AI methods in predictive maintenance. 
\end{itemize}

The rest of the paper is organized as follows. Section \ref{sec:related_work} provides an overview of the related work. Section \ref{sec:methodology} describes the methodology of the proposed framework. Section \ref{sec:experiments_results} provides a detailed description of the dataset, experiments, and experimental results. Finally, Section \ref{sec:conclusion_future_work} concludes the work. 

\section{Related Work}
\label{sec:related_work}
In this section, we provide an overview of the existing literature on both predictive maintenance and explainable AI. In the first part, we focus on the literature on predictive maintenance by highlighting some recent works in the domain. In the second part, we provide an overview of explainable AI techniques and key applications where AI could be beneficial. 

\subsection{Predictive Maintenance}
The literature reports several interesting works on predictive maintenance, where different aspects of predictive maintenance are explored \cite{jagatheesaperumal2021duo,afridi2021artificial}. Predictive maintenance generally involves three activities namely (i) data acquisition, (ii) data processing, and (iii) maintenance decision making, and the research in the domain mainly focuses on these areas \cite{selcuk2017predictive}. For data acquisition, the predictive maintenance techniques heavily depend on sensor technologies, which provide relevant and useful information on the machine conditions. Based on the nature of sensors, predictive maintenance techniques can be roughly divided into three categories including (i) existing sensor-based maintenance, (ii) test-sensor-based maintenance, and (iii) test signal-based maintenance techniques \cite{zonta2020predictive}.  

The literature also reports several interesting frameworks of intelligent data processing and handling for predictive maintenance. For instance, Shcherbakov et al. \cite{shcherbakov2020proactive} provides an overview of several data processing techniques and pipelines for data handling and processing for cyber-physical systems maintenance. Similarly, Yan et al. \cite{yan2017industrial} provide a detailed overview of the challenges associated with heterogeneous industrial data handling and processing for predictive maintenance. 

Predictive decision-making is one of the most explored topics in predictive maintenance. To this aim, several interesting solutions have been proposed over the years. The majority of the initial efforts in this direction rely on conventional/statistical ML algorithms, such as Random Forst, Support Vector Machines (SVMs), and decision trees \cite{afridi2021artificial}. For instance,
Kusiak et al. \cite{kusiak2011prediction} relied on two classical ML algorithms namely decision trees and SVMs, which were trained on feature vectors composed of 60 sensor readings, for the predictive maintenance of wind turbines. Other classical ML algorithms that are widely used for predictive maintenance include RF and Naive Bayes. These algorithms are normally trained raw sensor data or handcrafted features depending on the nature of the data \cite{afridi2021artificial}. A vast majority of the literature also relies on fuzzy logic and Hidden Markov Models (HMMs) for predictive maintenance. For instance, Zaki et al. \cite{zaki2019fault} and Omoregbee et al.  \cite{omoregbee2018fault} employed Fuzzy logic and HMMs for predictive maintenance of renewable energy systems, respectively. However, recently the trend shifted towards the use NNs, and the majority of the recently proposed solutions rely on different types of NNs, such as MLP, Convolutional Neural Networks (CNNs), and Long short-term memory (LSTM) \cite{afridi2021artificial}. The choice of these algorithms mainly depends on the nature of the data. For example, CNNs are mostly used for predictive maintenance using visual content \cite{shin2021ai,ulmer2020early}. LSTM-based solutions, on the other hand, are more effective for the analysis of sequential/time series data \cite{zou2021bearing,xiang2021fault}. 

\subsection{Explainable AI}
Over the last few years, explainable/interpretable AI got the attention of the research community. The literature reports several studies where it is demonstrated that in critical applications, such as healthcare, education, defense, and transportation, predictive capabilities of ML algorithms alone are not enough rather the algorithms should be interpretable \cite{ahmad2022developing,khan2021explainable}. Explainability/interpretability, which aims at highlighting the causes behind the AI models' predictions, could be obtained either by developing explainable AI algorithms or providing an explanation of the so-called black-box AI algorithms \cite{arrieta2020explainable}. However, there is a trade-off between accuracy/performance of AI algorithms and interpretation \cite{kamwa2011accuracy}. Therefore, a majority of the explainable AI frameworks rely on model agnostics methods for the interpretation of AI models \cite{ahmad2022developing,bennetot2021practical}. To this aim, several interesting techniques are proposed. Some most commonly used techniques include LIME \cite{ribeiro2016should}, SHAP \cite{lundberg2017unified}, Grad-CAM \cite{selvaraju2017grad}, and DiCE \cite{mothilal2020explaining}. 

Some key applications in which explainable AI has been widely explored include healthcare \cite{adadi2020explainable}, education \cite{tjoa2020survey}, security \cite{vigano2020explainable}, and other smart cities applications \cite{ahmad2022developing}. The applications of explainable AI have been recently also introduced in industry \cite{gade2019explainable}. However, most of the literature aims to analyze its applicability, challenges, and advantages in different industrial applications \cite{gade2019explainable}. For instance, Shukla et al. \cite{shukla2020opportunities} analyzed the opportunities of explainable AI in aerospace predictive maintenance. The authors also provide a detailed overview of the challenges associated with predictive maintenance in the domain. In contrast to most of the works reported in the literature, in this work, we propose an explainable regression framework for one of the most crucial applications of predictive maintenance namely the prediction of the remaining useful life of a machine.

\section{Methodology}
\label{sec:methodology}
\begin{figure*}[!h]
\centering
\includegraphics[width=.75\linewidth]{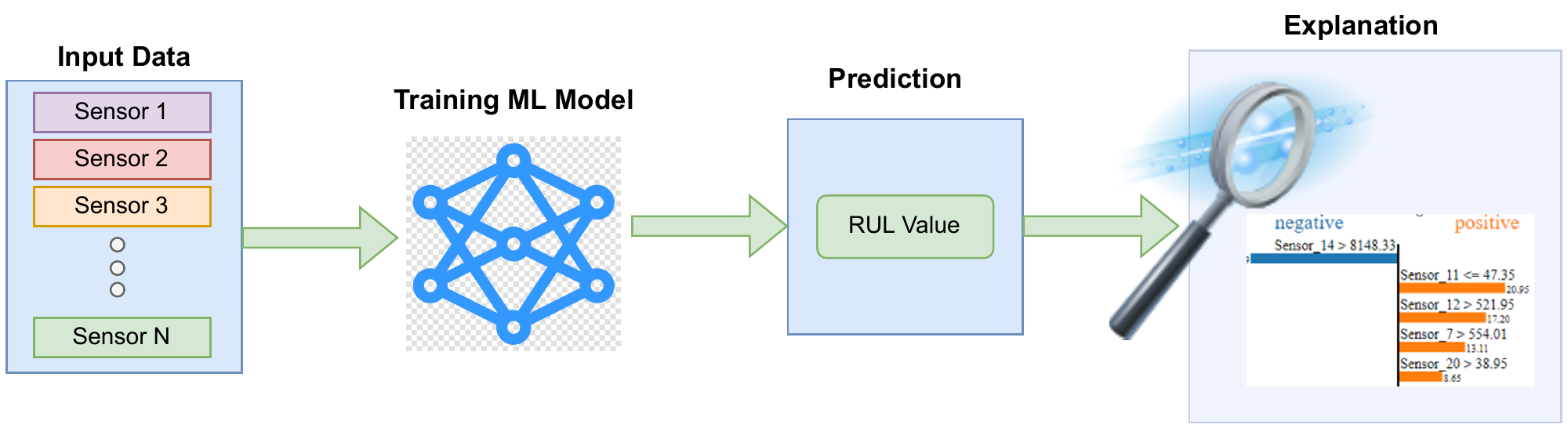}
\caption{Block diagram of the proposed methodology.}
	\label{fig:methodology}
\end{figure*}

Figure \ref{fig:methodology} provides the block diagram of the proposed explainable predictive maintenance framework. The framework is mainly composed of two components namely (i) features selection and ML-based prediction, and (ii) explanation of the predictions. In the first part, we rely on several ML algorithms for the prediction of the remaining useful life of the machine components. In the second part, two different algorithms are used for the explanation of the model's predictions. We note that the main contribution of the work lies in the explanation part. In the next subsections, we provide a detailed description of each of the phases.

\subsection{ML-based Prediction}
For the prediction of the useful remaining life of the machine components, we rely on several algorithms including (i) Random Forest (RF), (ii) ElasticNet with Generalized Linear Models (GLMs), (iii) Gradient Boosting, (iv) Support Vector Machines (SVMs), and a (v) Neural Network (NNs) model. A description of each of the methods is provided below. 

\begin{itemize}
    \item \textbf{RF-based Prediction}: RF is one of the most widely used methods. In this work, we use it as one of our baseline methods. In this approach, as a first step, a shallow RF model is used to identify more important/influential features by plotting a chart of feature ranking. After the plotting feature ranking, the less important features (i.e., sensors values) are dropped. An RF model is then trained on the selected features. Table \ref{tab:RF} provides the values of hyperparameters of the model used in the experiments.
    \item \textbf{SVM-based Prediction}: Our second baseline method is based on SVMs. SVMs are one of the most widely used algorithms for classification problems. However, it is rarely used for regression. It follows the same rules and criteria for regression tasks where the aim is to identify a function approximating the mapping from the input to real numbers based on training samples. One of the key processes in SVM-based prediction is the selection of hyperparameters values. To this aim, we rely on a grid-search algorithm to find the best combination of SVM hyperparameters.
    \item \textbf{Gardient Boosting-based Prediction}: Our third method is based on Gradient Boosting \cite{hastie2009elements}, which is also one of the most widely used algorithms for classification and regression tasks. It is an ensemble method where multiple learning algorithms, which are also called weak learners, are combined to obtain better predictive performance. In our case, the weak learners are based on decision trees. Similar to the other models, we rely on the grid-search approach for the selection of hyperparameters of the model.
    \item \textbf{ElasticNetGLM-based Prediction}: Elastic net regularization pairing with GLMs is one of the widely used regularization methods. It allows to filter out unimportant and highly correlated features and helps to improve the performance of the model. In this work, we use an ML library namely Scikit-learn for the implementation of the ElasticNetGLM model. For tuning the hyperparameters of the model, we rely on the grid-search algorithm that allows us to find the best combination of the hyperparameters. Table \ref{tab:ElasticNet} provides the summary of the parameters used in the model.

    \item \textbf{NNs-based Prediction}: Based on the proven performances in other applications, we also propose an NNs-based solution for the prediction of the useful remaining life of the machine components. To this aim, we propose an MLP regressor model. Our MLP regressor is composed of a total of 50 hidden layers, which are trained using backpropagation without an activation function in the output layer. Moreover, we used the square error as the loss function resulting in continuous values as an output.
\end{itemize}

\begin{table}[!h]
\caption{Parameters setting of the FR model.}
		\label{tab:RF} 
\begin{tabular}{|c|c|}
\hline
\textbf{Attribute} & \textbf{Value} \\ \hline
	max-depth & 9  \\ \hline
   max-features & auto  \\ \hline
  min-samples-leaf & 10   \\ \hline
    min-samples-split & 2  \\ \hline
   n-estimators & 10  \\ \hline
\end{tabular}
\end{table}

\begin{table}[!h]
\caption{Parameters setting of the ElasticNetGLM model.}
		\label{tab:ElasticNet} 
\begin{tabular}{|c|c|}
\hline
\textbf{Attribute} & \textbf{Value} \\ \hline
Alpha	  & 0.01   \\ \hline
  l1-ratio  & 0.01   \\ \hline
  copy-X & True  \\ \hline
    fit-intercept  & True	   \\ \hline
      selection & Cyclic   \\ \hline
        tol & True	 0.0001   \\ \hline
\end{tabular}
\end{table}

\begin{table}[!h]
\caption{Parameters setting of the MLP model.}
		\label{tab:MLP} 
\begin{tabular}{|c|c|}
\hline
\textbf{Attribute} & \textbf{Value} \\ \hline
	Model  &  MLP Regressor \\ \hline
    Hidden Layers &  50 \\ \hline
  Max Iteration	  &  1000 \\ \hline
  Learning Rate   &  Adaptive \\ \hline
\end{tabular}
\end{table}
\subsection{Model's Explanation}
For the models' explanation, we mainly rely on two methods namely (i) LIME, and (ii) SHAP. In the next subsections, we provide a detailed description of each of the methods.
\subsubsection{\textbf{LIME-Local Interpretable Model-Agnostic Explanations}}
LIME \cite{ribeiro2016should} is one of the most commonly used methods for the interpretation of ML models' predictions. One of the key advantages of LIME is that it is a model agnostics method and could be used for the explanation/interpretation of any model. For the explanation of an ML model, LIME perturbs the input samples and analyzes the changes in the prediction of the model. This simple working mechanism makes it a preferable choice for model interpretation compared to model-specific methods, which require a deeper understanding of the underlying models. 

LIME provides local interpretation, which means the model's behavior is described by analyzing the response of a model to changes in a single data sample. Here the intuition is to analyze causing behind a particular prediction by answering questions like ''why was this prediction made?'' or ''which features caused the prediction?''. It produces results in the form of a list of explanations highlighting the contribution of the individual feature as detailed in Section \ref{sec:experiments_results}. We note that the idea and working mechanism of LIME is different from a related concept of ''\textit{feature importance}'', which is generally conducted over the entire datasets.

LIME could be used for the explanations of models deployed in different application domains including textual, tabular (i.e., sensor data), and visual content. The literature already reports the effectiveness of the method in several interesting human-centric applications, such as healthcare and other smart cities applications \cite{ahmad2022developing}. 

\subsubsection{\textbf{SHAPE-Shapley Additive Explanations}} 
Our second explanation approach is based on another state-of-the-art technique namely SHAPE. The method was introduced by Lundberg et al. \cite{lundberg2017unified} to cope with the limitations of the existing methods. Similar to LIME, SHAP provides explanations of individual predictions and could be used for the explanation of any model. However, in contrast to LIME, SHAP provides both local and global explanations. The main difference between global and local explanations lies in the level/scope of explanation. The global interpretations/explanations include complete insights into the general/overall behavior of the model. The local explanations/interpretations describe the causes behind a decision on an individual data sample.

Similar to LIME, SHAP could be used for the explanation of ML models trained on different types of data including textual, visual, and tabular data. The literature reports the effectiveness of the method in several application domains, such as healthcare, defense, agriculture, etc \cite{ahmad2022developing,zhou2022identification,elshawi2021interpretability}. 

\section{Experiments and Results}
\label{sec:experiments_results}

\subsection{Dataset}
\label{sec:dataset}
For the evaluation of proposed solutions, we used a dataset composed of the engine degradation simulation (C-MAPSS) data, which is collected in a simulated engine degradation environment under different combinations of operational conditions and modes \cite{saxena2008damage}. The dataset is provided by the Prognostics Center of Excellence (PCoE) where the data is based on time series ranging from the working state to the failure state of the components. The dataset provides a 24 features vector containing 21 sensor readings and 3 operational settings. The sensor readings are mostly related to temperature, pressure, the fan speed of an engine, fuel, etc. The description of all of these sensor readings is provided in Table \ref{tab:sensor}. 

The dataset is widely used for fault detection and prognostics (i.e., predicting the time at which the machine components will no longer work). In this paper, we are interested in predicting the Remaining Useful Life (RUL) of machine components, which is a continuous target/value. We treat the problem as a regression task. The dataset is composed of more than 20,000 data samples, which are divided into training and test sets. The training set is composed of 16504 samples while the test set contains  4127 samples.
\begin{table}[!h]
\caption{Description of the sensor readings provided in the dataset.}
		\label{tab:sensor}
		\scalebox{.75}{
\begin{tabular}{|c|c|c|c|}
\hline
\textbf{Sensor No} & \textbf{Readings} & \textbf{Sensor No} & \textbf{Readings} \\ \hline
1 & Total temperature at fan inlet & 2 & Total temperature at LPC outlet \\ \hline
3 & Total temperature at HPC outlet & 4 & Total temperature at LPT outlet \\ \hline
5 & Pressure at fan inlet & 6 & Total pressure in bypass-duct \\ \hline
 7 & Total pressure at HPC outlet & 8 & Physical fan speed \\ \hline
 9 & Physical core speed & 10 & Engine pressure ratio \\ \hline
 11 & Engine pressure ratio & 12 & Ratio of fuel flow to Ps30 \\ \hline
 13 & Corrected fan speed & 14 & Corrected core speed \\ \hline
 15 & Bypass Ratio & 16 & Burner fuel-air ratio \\ \hline
 17 & Bleed Enthalpy & 18 &  Demanded fan speed\\ \hline
 19 & Demanded corrected fan speed & 20 &HPT coolant bleed  \\ \hline
 21 & LPT coolant bleed & - & - \\ \hline
\end{tabular}}
\end{table}

\subsection{Experimental Results}
\label{sec:results}
In this section, we provide the experimental results of the proposed work. Firstly, we report the results of all the algorithms employed in this work followed by some samples of the explanations produced by LIME and SHAP.

\subsubsection{\textbf{Prediction Results}}
Table \ref{tab:results} provides the experimental results of all the algorithms employed in this work including the classical and NNs-based solutions in terms of Mean Squared Error (MSE) and Mean Absolute Error (MAE) on a test set containing a total of 4127 samples. We note that there is a trade-off between performance and explainability. The classical ML algorithms are more explainable compared to NNs, however, their performance is generally on the lower side. Similar trends have been also observed in this work. As can be seen, the MSE and MAE values for the NNs-based solution are significantly lower compared to the classical algorithms, such as RF and SVMs. 

As far as the explanations of the NNs-based solutions is concerned, this could be overcome with the use of model agnostic methods of explainability. Next, we analyze the explanations provided by two model agnostic explainable AI methods.

\begin{table}[]
\center
\caption{Experimental results in terms of Mean Squared Error (MSE) and Mean Absolute Error (MAE) on a test set containing 4127 samples.}
		\label{tab:results}
\begin{tabular}{|c|c|c|}
\hline
\textbf{Method} & \textbf{MSE} & \textbf{MAE} \\ \hline
RF & 1767.06 & 29.84 \\ \hline
ElasticNetGLM & 2043.03 & 34.60 \\ \hline
 Gradient Boosting&  1768.45 &  29.92\\ \hline
 SVMs&  2043.03& 34.60 \\ \hline
 MLP & 1742.08 & 28.46 \\ \hline
\end{tabular}
\end{table}

\subsubsection{Model Explanations via LIME and SHAP}
\begin{figure*}[!h]
\centering
\includegraphics[width=.85\linewidth]{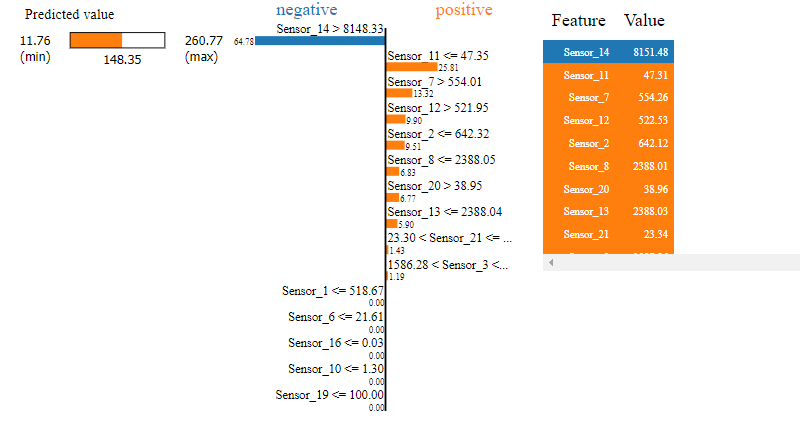}
\caption{Sample explanations provided by LIME. Here the actual value of RUL of the component is 151 while the predicted value is 148.3.}
	\label{fig:LIME}
\end{figure*}

\begin{figure*}[!h]
\centering
\includegraphics[width=.85\linewidth]{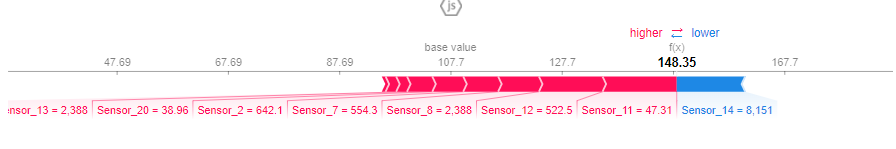}
\caption{Sample explanations provided by SHAP. Here the actual value of RUL of the component is 151 while the predicted value is 148.3.}
	\label{fig:SHAP}
\end{figure*}

Figure \ref{fig:LIME} and Figure \ref{fig:SHAP} provides explanations generated by LIME and SHAP method, respectively. In this sample, the actual value of RUL of the machine component is 151 cycles while the predicted value is 148.35 cycles. 

In Figure \ref{fig:LIME}, the features (i.e., sensor values) on the right side in red color are the ones that contribute to increasing the prediction values while the ones in blue color are the features that have a negative effect or decrease the predicted value. For example, the values of the sensors 11, 7, 2, etc. on the right side indicate that the machine component is in a good condition and its RUL is supposed to be high. The sensor value on the left side in blue color, for example, $sensor-1 =< 518.67$ indicates that the conditions of the component are not good and it tries to reduce the predicted RUL value.

Figure \ref{fig:SHAP} provides the explanation of the same sample. Similar to LIME, SHAP explanations are also composed of several values including:

\begin{itemize}
    \item The predicted value (i.e., 148.35)
    \item The base value, which is the average of the model output over the training dataset.
    \item The feature values that contributed to the prediction. The red values are the feature values that increased the predicted value while the values in blue color (i.e., $sensor-14 = 8.151$ contributed to the reduction of the predicted RUL value. In other words, the sensor values in red color mean the condition of the component is good while the blue ones indicate something is wrong with the component.
    \item The size/length of the arrow shows the impact of the feature on the prediction. For example, in the given sample, sensor-11, sensor-12, and sensor-14 have a higher impact on the prediction.
\end{itemize}

\section{Conclusions and Future Work}
\label{sec:conclusion_future_work}
In this work, we proposed an explainable regression framework for the prediction of the RUL of machine components. We evaluated several ML techniques including classical ML and NNs approaches for the prediction. For the explanation of the models, we employed two different models agnostic explainable AI methods. The explanation provided by these methods is very insightful that could help the stakeholders in making correct decisions in such critical applications. The explanation of the models could also help the developers to rectify the limitations of their proposed solutions. In the future, we aim to further extend the scope of the work by tackling more relevant tasks of predictive maintenance.


\ifCLASSOPTIONcaptionsoff
 \newpage
\fi

\bibliographystyle{IEEEtran}
\bibliography{biblo}

\end{document}